\documentclass{article}

     \usepackage[preprint,nonatbib]{neurips_2021}

\usepackage[utf8]{inputenc} % allow utf-8 input
\usepackage[T1]{fontenc}    % use 8-bit T1 fonts
\usepackage{url}            % simple URL typesetting
\usepackage{booktabs}       % professional-quality tables
\usepackage{amsfonts}       % blackboard math symbols
\usepackage{nicefrac}       % compact symbols for 1/2, etc.
\usepackage{microtype}      % microtypography
\usepackage{xcolor}         % colors

\usepackage{graphicx}
\usepackage{tikz}
\usepackage{comment}
\usepackage{amsmath,amssymb} % define this before the line numbering.
\usepackage{color}
\usepackage{graphicx}
\usepackage{amsmath}
\usepackage{amssymb}
\usepackage{tabularx}
\usepackage{pifont}
\usepackage{multirow}
\usepackage{subcaption}
\usepackage{arydshln}
\definecolor{citecolor}{HTML}{2980b9}
\definecolor{linkcolor}{HTML}{c0392b}
\usepackage[pagebackref=true,breaklinks=true,colorlinks,bookmarks=false,citecolor=citecolor,linkcolor=linkcolor]{hyperref}

\definecolor{cl1}{rgb}{0.23,0.47, 0.84}
\definecolor{cl2}{rgb}{0.90,0.56, 0.21}

\usepackage[accsupp]{axessibility}  % Improves PDF readability for those with disabilities.

\newcommand{\env}{\mbox{A-OKVQA}}

\title{\env: A Benchmark for Visual Question Answering using World Knowledge}

\author{
Dustin Schwenk$^1$, Apoorv Khandelwal$^1$, Christopher Clark$^1$ \\ \textbf{Kenneth Marino}$^2$, \textbf{Roozbeh Mottaghi}$^1$ \\
$^1$ PRIOR @ Allen Institute for AI\\
$^2$ Carnegie Mellon University
}

\begin{document}

\maketitle

\begin{abstract}
The Visual Question Answering (VQA) task aspires to provide a meaningful testbed for the development of AI models that can jointly reason over visual and natural language inputs. Despite a proliferation of VQA datasets, this goal is hindered by a set of common limitations. These include a reliance on relatively simplistic questions that are repetitive in both concepts and linguistic structure, little world knowledge needed outside of the paired image, and limited reasoning required to arrive at the correct answer. We introduce \env, a crowdsourced dataset composed of a diverse set of about 25K questions requiring a broad base of commonsense and world knowledge to answer. In contrast to the existing knowledge-based VQA datasets, the questions generally cannot be answered by simply querying a knowledge base, and instead require some form of commonsense reasoning about the scene depicted in the image.  We demonstrate the potential of this new dataset through a detailed analysis of its contents and baseline performance measurements over a variety of state-of-the-art vision--language models. \\

\centerline{\large{\url{http://a-okvqa.allenai.org/}}}

\end{abstract}

\section{Introduction}
The original conception of the Visual Question Answering (VQA) problem was as a Visual Turing Test~\cite{Geman2015VisualTT}. Can we give a computer an image and expect it to answer any question we ask to fool us into thinking it is a human? To truly solve this Turing Test, the computer would need to mimic several human capabilities including: visual recognition in the wild, language understanding, basic reasoning capabilities and a background knowledge about the world. Since the VQA problem was formulated, many of these aspects have been studied. Early datasets mostly studied the perception and language understanding problem on natural image datasets~\cite{antol15,malinowski14b,goyal2017making}. Other datasets studied complex chains of reasoning about procedurally generated images~\cite{johnson17a}. More recently, datasets include questions which require factual~\cite{marino19cvpr,wang17a,wang17b} or commonsense knowledge~\cite{Zellers2019FromRT}. 

But, to a large extent, VQA has been a victim of its own success. With the advent of large-scale pre-training of vision and language models~\cite{Zhang2021VinVLRV,Yang2021AnES,lu19vilbert,lu202012,Devlin2019BERTPO,radford2019language,brown2020language} and other breakthroughs in multi-modal architectures, much of the low-hanging fruit in the field has been plucked and many of the benchmark datasets have seen saturated performance. Even performance on the newer knowledge-based datasets has been improved by such models~\cite{Zhang2021VinVLRV}. So how can we continue developing yet more challenging datasets? What human capabilities are not yet expressed by current models?

\begin{figure}[tp]
    \centering
    \includegraphics[width=33pc]{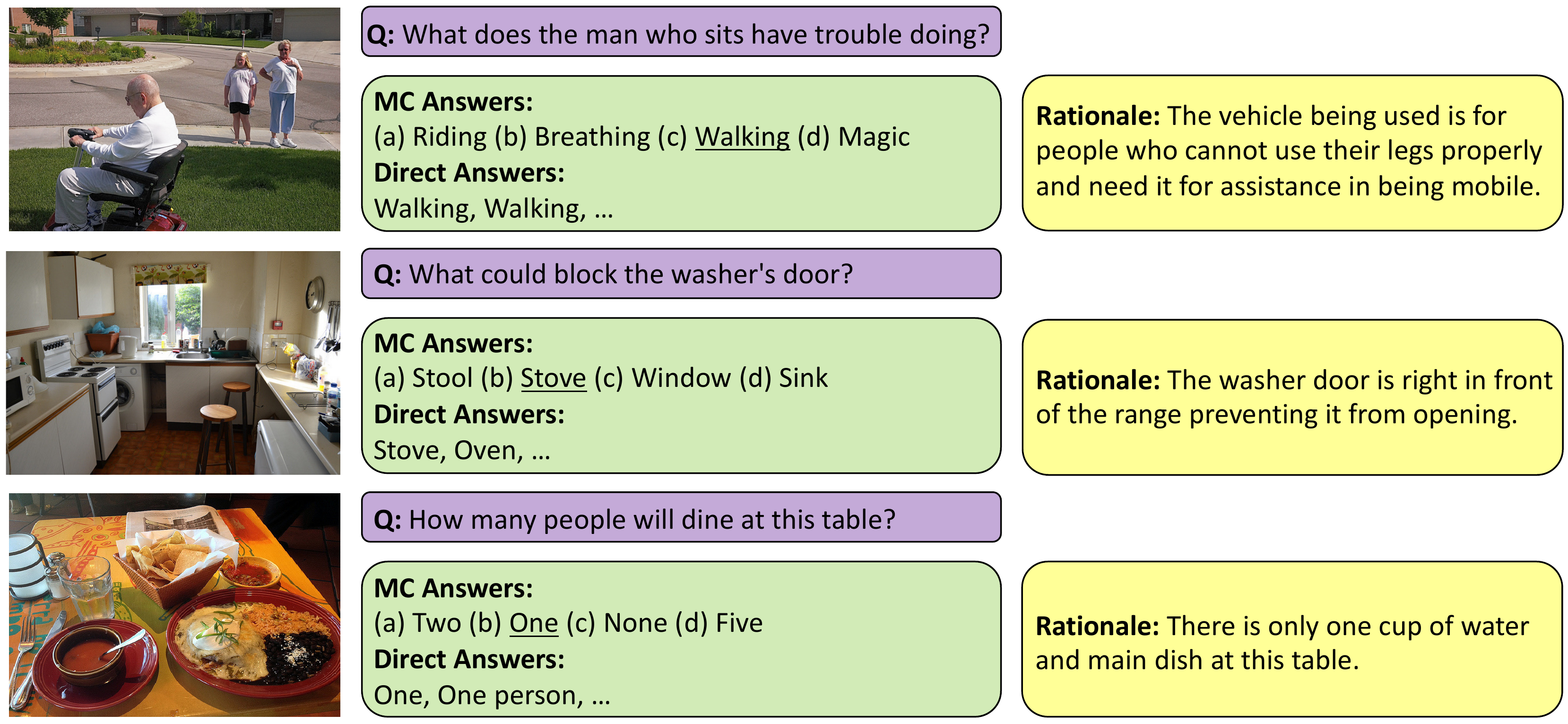}
    \caption{\textbf{\env\ dataset.} The dataset includes questions that require reasoning using a variety of knowledge types such as commonsense, world knowledge and visual knowledge. We provide Multiple-Choice (MC) as well as Direct Answer evaluation settings. There is a rationale associated to each question in the train set providing the explanation/knowledge for answering the question.}
    \label{figures:example_questions} 
\end{figure}

We propose the following. First, continuing the direction of past work in knowledge-requiring VQA, we further expand the areas of knowledge required. Our dataset requires diverse forms of outside knowledge including explicit fact-based knowledge that is likely to be contained in knowledge bases, commonsense knowledge about human social behavior, knowledge about the physics of the world, and visual knowledge. Not only do we expand the variety of knowledge our agent needs, but we also increase the complexity of reasoning systems needed to answer questions. We need models to recognize the image, understand the question, recall relevant knowledge, and use reasoning to arrive at an answer. For instance, in the first question shown in Figure~\ref{figures:example_questions}, the model should reason that people use that type of cart to avoid walking. Therefore, the old man should have trouble walking. In general, our dataset requires additional types of world knowledge compared to our previous work OK-VQA~\cite{marino19cvpr}. Hence, we call it Augmented OK-VQA (\env).

\env\ is composed of about 25K questions paired with both multiple choice (MC) answer options and ten free-form answers to allow for direct answer (DA) evaluation. The MC component of the dataset bypasses many difficulties inherent in direct answer evaluation and allows for a simple, clean accuracy score. This is particularly helpful given the greater variety in answers in \env\ questions. At the same time, we believe direct answer evaluation is important to encourage models with more real-world applicability. In addition to the questions and answers, we provide \emph{rationales} for each question. This is to allow for models to use this extra annotation to train reasoning or knowledge retrieval methods or to build more explainable VQA models. The rationales also validate that both reasoning and knowledge are required for answering questions in the dataset.

In this work, our contributions are: (i) A new benchmark VQA dataset requiring diverse sources of outside knowledge and reasoning; (ii) A detailed analysis of the dataset that highlights its diversity and difficulty; (iii) An evaluation of a variety of recent baseline approaches in the context of the challenging questions in \env; (iv) An extensive analysis of the results leading to interesting findings (e.g., how well models perform when answers are in the tail of the distribution, and also the complementarity of the studied models). 
We will release this dataset publicly.

\section{Related Work}
\label{sec:relatedwork}

\textbf{Visual Question Answering.}
Visual Question Answering (VQA) has been a common and popular form of vision and language reasoning. Many datasets on this task have been proposed~\cite{malinowski14b,antol15,gao15,yu15,ren15,zhu16,tapaswi16,krishna17} but most of these do not require much outside knowledge or reasoning, often focusing on recognition tasks such as classification, attribute detection and counting.

\textbf{Knowledge-based VQA datasets.} Several previous works have studied the problem of knowledge-based VQA. The earliest explicitly knowledge-based VQA datasets were KB-VQA~\cite{wang17a} and FVQA~\cite{wang17b}. While these benchmarks did specifically require knowledge for questions, the knowledge required for these benchmarks is completely ``closed''. FVQA~\cite{wang17b} is annotated by selecting a triplet from a fixed knowledge graph. This forces the questions to require knowledge, but because the question is written based on this knowledge, these questions are fairly trivial once the knowledge is known and do not require much reasoning. In addition, the knowledge required is explicitly closed to the knowledge graphs used to generate the dataset, so these datasets can only test knowledge retrieval on those specific graphs. KVQA~\cite{shah2019kvqa} is based on images in Wikipedia articles. Because of the source of the images, these questions tend to mostly test recognizing specific named entities (e.g., Barrack Obama) and then retrieving Wikipedia knowledge about that entity rather than commonsense knowledge.

Most similar to our work is OK-VQA~\cite{marino19cvpr}. This dataset was an improvement over prior work in terms of scale, and the quality of questions and images. It also has the property that the required knowledge was not ``closed'' or explicitly drawn from a particular source, and could be called ``open''-domain knowledge. While this is an improvement over the previous works, it still suffers from problems which we address in this work. The knowledge required, while ``open'' is still biased towards simple lookup knowledge (e.g., what is the capital of this country?) and most questions do not require much reasoning. In contrast, our dataset is explicitly drawn to rely on more common-sense knowledge and to require more reasoning to solve. In addition, our dataset includes ``rationale'' annotations, which allow knowledge-based VQA systems to more densely annotate their knowledge acquisition and reasoning capabilities.
S3VQA \cite{Jain2021SelectSS} analyzes OK-VQA and creates a new dataset which includes questions that require detecting an object in the image, replacing the question with the word for that object and then querying the web to find the answer. Like OK-VQA, it even more explicitly has the problem of questions usually requiring a single retrieval rather than much commonsense knowledge or reasoning.

Another related line of work is Visual Commonsense Reasoning (VCR)~\cite{Zellers2019FromRT} and VisualCOMET~\cite{Park2020VisualCOMETRA}. VCR is also a VQA dataset, but is collected from movie scenes and is quite focused on humans and their intentions (e.g. ``why is [PERSON2] doing this''), whereas our dataset considers questions and knowledge about a variety of objects. Similarly, VisualCOMET tests commonsense language and vision models on a movie dataset, but its expected output is a scene graph for the image (e.g., ``After, [PERSON] is likely to''). Additionally, the Ads Dataset~\cite{Hussain2017AutomaticUO} is a dataset requiring knowledge about the topic and sentiments of the ads. Other datasets have considered knowledge-based question answering for a sitcom~\cite{Garca2020KnowITVA} and by using web queries~\cite{Chang2021WebQAMA}. 

\textbf{Explanation / Reasoning VQA.} Visual reasoning on its own has been studied in several VQA datasets. In CLEVR~\cite{johnson17a}, the image and question are automatically generated from templates and explicitly require models to go through multiple steps of reasoning to correctly answer. This dataset and similar datasets which rely on simulated images suffer from lack of visual realism and lack of richness in the images and questions and are thus prone to be overfit to with methods achieving nearly 100\% accuracy~\cite{Yi2018NeuralSymbolicVD}. Our dataset requires reasoning on real images and free-form language. Other works~\cite{park2018multimodal,Li2018TellandAnswerTE} have collected or extracted justifications on the VQAv2~\cite{goyal2017making} dataset. However, VQAv2 mostly focuses on questions about object attributes, counting and activities, which do not require reasoning on outside knowledge. 

\textbf{Knowledge / Commonsense in NLP.}
Question answering with knowledge and commonsense is also a well-studied problem in natural language processing. This takes the form of knowledge base completion~\cite{bordes2011learning}, knowledge-based question answering~\cite{rajpurkar16squad} to open-domain question answering~\cite{chen2017reading}. \cite{berant13,yao14,bordes14} address question answering from specific knowledge sources. This includes open-domain question answering~\cite{chen2017reading,wang2017r,yang2015wikiqa,yang2019end,Sun2018OpenDQ} and question answering from Wikipedia SQu--AD \cite{rajpurkar16squad,Rajpurkar18}. Much work has also been done in commonsense question answering as in CommonsenseQA~\cite{Talmor2019CommonsenseQAAQ}, where there is no direct source of knowledge but the agent must have general ``commonsense'' to answer the question. 
\section{\env\ Collection}

\textbf{Image source.} The first requirement of an image source for this knowledge-based VQA task is that it has an abundance of visually rich and interesting images. Images containing a small number of objects are typically quite challenging to write questions requiring outside knowledge to answer. We used images from the 2017 partitioning of the COCO dataset~\cite{Lin2014MicrosoftCC} in the creation of \env\ because: (1) it has many images cluttered with multiple objects and entity types, (2) it is an established dataset with many associated models already in existence. To ensure suitable images for annotation, we do some additional filtering to remove uninteresting images: For the training and validation sets, we define images with more than three objects as ``interesting'' and select those for question writing. For the test set, which lacks object annotation, we train a ResNet-50 classifier to distinguish ``interesting'' images based on this criteria, achieving an accuracy of $\textbf{78\%}$ on the validation set. After multiple rounds of filtering (described below), we obtain 23.7K unique images.

\textbf{Question collection \& filtering.}
The questions in \env\ were written and refined over several rounds of annotation by 437 crowd-workers on the Amazon Mechanical Turk platform and refined through several manual and automated filtering steps to increase overall quality. As a first quality assurance measure, workers completed a qualification task to demonstrate their ability to write questions that met our criteria, namely that questions require: (1) looking at the image to answer, (2) some commonsense or specialized knowledge, (3) some thinking beyond merely recognizing an object, and (4) not be too similar to previous questions. 

To help ensure the last point, we clustered images by CLIP~\cite{Radford2021LearningTV} visual features and batched similar images together so that the same worker wrote questions sequentially for related images (e.g., a worker might write questions for several images showing baseball games in one task) to cut down on repetitive questions. As an added measure to encourage question diversity, we maintained a database of questions written and required users to check a new question against these by displaying the five previous questions most similar in terms of their RoBERTa~\cite{Liu2019RoBERTaAR} embeddings. We used Pythia~\cite{singh2019TowardsVM} pre-trained on VQAv2 as a first automated check for questions we considered trivial, removing any question for which the model predicted the correct answer choice. Next, questions were screened by three other workers and only included if the majority agreed that it met our criteria for inclusion. In all, 37,687 questions, or $\textbf{60\%}$ of post-qualification questions were excluded from the dataset by this process. After questions and their multiple choice options were complete, nine additional free-form answers were collected for each question by a separate pool of workers.

\textbf{Rationales.} After questions and multiple-choice answer options were collected and validated, we initiated a task to collect rationales. Workers were given a question and answer options and asked to explain in one to two simple sentences why a particular answer was correct, including any necessary facts or knowledge about the world not contained in the images. Workers were given examples and went through a qualification process to assure high-quality output. For each question, we collected three rationales.

\section{Dataset Statistics}

\label{sec:dataset_stats}

\textbf{Question/Answer/Rationale statistics.}
After all rounds of annotation, the \env\ dataset contains 24,903 Question/Answer/Rationale triplets, split into 17.1K/1.1K/6.7K for train, validation and test. These preserve the COCO 2017 train/val/test splits. The average length of the questions, answers, and rationales, and the number of their unique words are shown in Table~\ref{tab:comparison}.

\begin{figure}[tp]
     \centering
     \begin{subfigure}[b]{0.45\textwidth}
         \centering
         \includegraphics[width=0.9\textwidth]{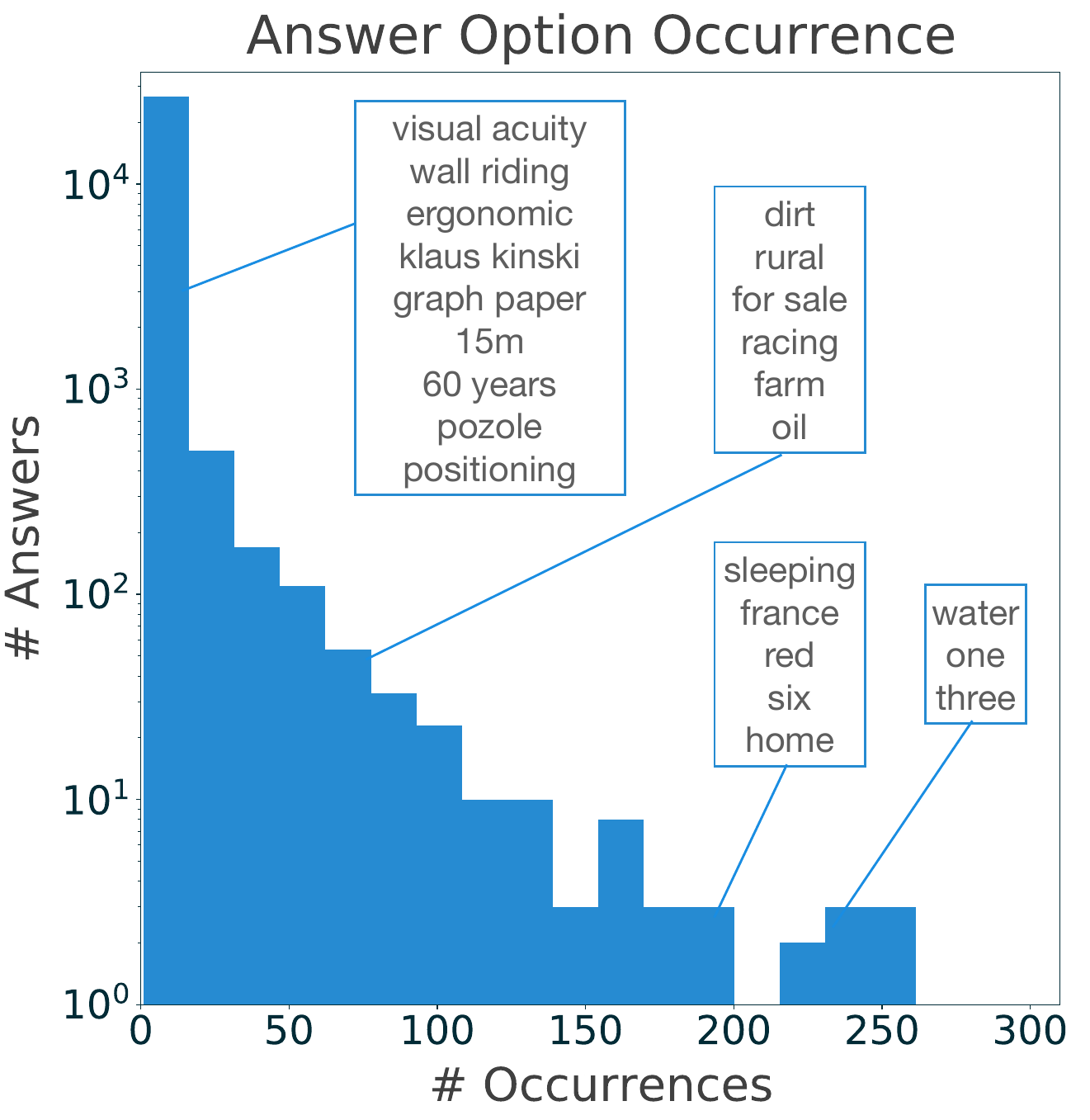}
         \caption{Answer occurrence distribution of \env.}
         \label{figures:answer_option_freq} 
     \end{subfigure}
     \hfill
     \begin{subfigure}[b]{0.45\textwidth}
         \centering
         \includegraphics[width=0.9\textwidth]{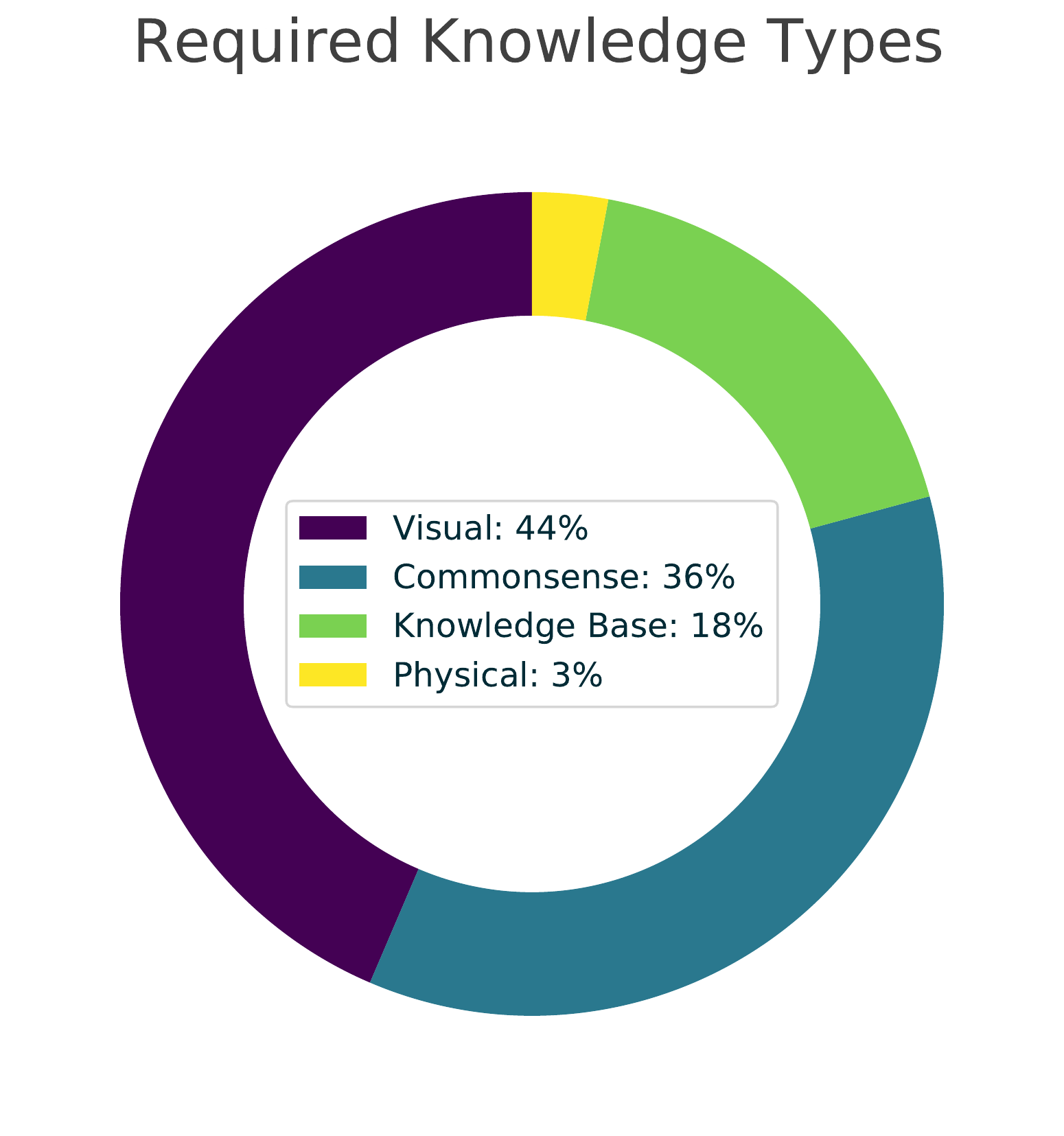}
         \caption{Knowledge type distribution for a random subset of the dataset.}
         \label{figures:knowledge_type_chart} 
     \end{subfigure}
     \caption{\textbf{Dataset statistics.}}
     \label{fig:stat}
\end{figure}

In Figure~\ref{figures:answer_option_freq} we show the distribution of answer options in our dataset. What we see is a fairly typical long-tail distribution of labels, as is seen in many open-labeled image tasks~\cite{Zhu14}. A few answers occur quite often in the dataset, but overall, most answers in the dataset fall into the long tail of the distribution.

We are also interested to know the amount of overlap in the answer set between the train set and val and test sets. We find that in the val set, the ground-truth answer for $87.6\%$ of questions appears in the train set while in the test set $82.7\%$ do. This shows that there is indeed some reasonable similarity between the sets, but also that a significant portion of the held out sets require an answer that the model will not have seen during training, requiring the model to be able to generate out-of-distribution answers or generate answers based on some knowledge outside of the dataset.

\textbf{Comparison with other datasets.} In Table~\ref{tab:comparison} we show dataset properties and statistics for \env \ compared to related datasets. We see that compared to the more knowledge-focused natural image datasets such as OK-VQA, we have between 2-10x more questions while VCR (focused on images of people in movies) has about 10x as ours. This is unsurprising because we intentionally filter similar questions, making our questions more diverse (see Table~\ref{table:BERTDist}), but difficult to collect at scale. Our dataset has annotations for both multiple choice and direct answer evaluation. Our dataset also has rationales, unlike OK-VQA, S3VQA and KB-VQA. FVQA has knowledge tuples as rationales rather than full sentences. Most similar to our rationales is VCR. Unlike all of these, we collect 3 rationales rather than just 1 as the rationales are more knowledge-based and have more possible variation within the same question. Our average question lengths are long compared to many of these datasets except for S3VQA which has the longest and VCR which is about on par. Ours also contains the most unique words except for VCR, likely because that dataset has more questions.

\begin{table*}[tp]
\centering
\scriptsize
\begin{tabular}{l | c c c c c c c}
    \toprule
    & \multirow{2}{*}{\parbox{0.3cm} {\centering Q}} & 
    \multirow{2}{*}{\parbox{0.1cm} {\centering I}} & 
    \multirow{2}{*}{\parbox{0.9cm} {\centering Rationale}} &
    \multirow{2}{*}{\shortstack{Knowledge\\type}} &
    \multirow{2}{*}{Ans type} & 
    \multirow{2}{*}{\shortstack{Avg. length\\(Q/A/R)}} &
    \multirow{2}{*}{\shortstack{unique words\\(Q/A/R)}}\\\\
    \midrule
      KB-VQA~\cite{wang17a} & 2,402 & 700 & \ding{56} & fixed KB & DA & 6.8/2.0/NA & 530/1,296/NA \\
      FVQA~\cite{wang17b} & 5,826 & 2,190 & \ding{51} & fixed KB & DA & 9.5/1.2/NA & 3,010/1,287/NA\\
      OK-VQA~\cite{marino19cvpr} & 14,055 & 14,031 & \ding{56} & factoid & DA & 8.1/1.3/NA &  5,703/11,125/NA\\
      S3VQA~\cite{Jain2021SelectSS} & 7,515 & 7,515 & \ding{56} & factoid & DA & 12.7/2.8/NA & 7,515/8,301/NA\\
      VCR~\cite{Zellers2019FromRT} & 290k & 99,904 & \ding{51} & people actions & MC & 8.7/7.7/16.8 & 11,254/18,861/28,751\\
      %VCR~\cite{Zellers2019FromRT} & 290k & 99,904 & \ding{51} & people actions & MC & 8.7/7.7/16.8 & 11k/18k/28k\\
      \midrule
      \textbf{\env} & 24,903 & 23,692 &  \ding{51} & common/world & DA/MC & 8.8/1.3/11.0 & 7,248/17,683/20,629\\
\bottomrule
\end{tabular}
\caption{\textbf{Comparison of various knowledge-based VQA datasets.} Data based on publicly reported numbers and/or our analysis of publicly available annotations (therefore some answer statistics may exclude test sets). Answer statistics for \env\ based on the direct answer set. Q: question, I: image, A: answer, R: rationale, DA: Direct Answer, MC: Multiple Choice. NA indicates lack of rationales or, for FVQA are KB triplets, so we do not compare the lengths.}
\label{tab:comparison}
\end{table*}

\textbf{Knowledge types.} The most significant factor differentiating our dataset is the kind of knowledge required. Datasets such as FVQA have fixed knowledge bases that are used to write the questions, and so the knowledge required can be found in e.g. ConceptNet~\cite{liu2004conceptnet} directly. OK-VQA and S3VQA focus on more factoid knowledge (e.g., years of invention or countries of origin). In S3VQA in particular, researchers found that these datasets take the form of finding an entity in the image and/or question and searching and retrieving knowledge about that particular entity (see~\cite{Jain2021SelectSS}). VCR is overwhelmingly images of people interacting in television shows and movies and requires images to have people in them. Thus, the required knowledge is very focused on commonsense about human behavior and intentions. In our dataset, we require broader areas of knowledge including the factoid knowledge likely to be contained in knowledge bases (as in FVQA, KBVQA, OKVQA and S3VQA), and commonsense knowledge (like VCR but broader than just about people). 

To analyze the knowledge required in \env\ more quantitatively, we annotated a randomly sampled subset of 1,000 questions in the \env\ test. In this experiment, we ask the annotators to label what kind of knowledge type was required to answer the questions. The choices were: (1) \textbf{Commonsense} knowledge about human social behavior (e.g., that many donuts being made in a cart implies they are for sale rather than for personal consumption), (2) \textbf{Visual knowledge} (e.g., muted color pallets are associated with the 1950s), (3) \textbf{Knowledge bases} (e.g., hot dogs were invented in Austria), (4) \textbf{Physical knowledge} about the world that humans learn from their everyday experiences (e.g., shaded areas have a lower temperature than other areas). The distribution is shown in Figure~\ref{figures:knowledge_type_chart}. Most of our questions focus around commonsense and visual knowledge. It should be noted that sometimes there is no clear distinction between these two categories, and a question can belong to either category. 

\textbf{Question diversity.} To analyze the diversity of \env\ compared to other datasets, as a proxy, we use the average pairwise cosine distance between the questions in the dataset. We run our questions through a sentence transformer\footnote{Specifically multi-qa-MiniLM-L6-cos-v1~\cite{hgf} to avoid overlap with RoBERTa.} and compute the cosine distance between all pairs in the dataset. We then take the average of these. We see from Table~\ref{table:BERTDist} that our dataset has the most diversity on this metric. In particular, we see a large difference compared to VCR which has many similar questions such as ``What is going to happen next?'' and questions relating to what specific people in the scene are doing and why. We also compare the diversity of rationales to VCR and VQAv2 (using rationales from VQA-X~\cite{park2018multimodal} rationales). We also find that our rationales are much more diverse than in these datasets. Qualitatively, we also find that our dataset tends to have much more varied questions because it is taken from the more visually diverse COCO dataset (a quality shared by OK-VQA and VQAv2 which do almost as well on this metric) and requires more diverse kinds of knowledge.

\begin{table}[tp]
 \scriptsize
 \setlength{\tabcolsep}{5pt}
     \begin{center}
        \begin{tabular}{l | c  c}
        \toprule
      Dataset & Mean Q distance & Mean rationale distance\\ \midrule 
      FVQA~\cite{wang17b} & 0.6199 & \ding{55} \\      
      VCR~\cite{Zellers2019FromRT} & 0.7095 & 0.8017 \\       
      KB-VQA~\cite{wang17a} & 0.7192 & \ding{55} \\
      S3VQA~\cite{Jain2021SelectSS} & 0.8050 & \ding{55} \\       
      VQAv2~\cite{goyal2017making} & 0.8405 & 0.8228 \\      
      OK-VQA~\cite{marino19cvpr} & 0.8428 & \ding{55} \\
      \midrule
      \env & 0.8564 & 0.8779 \\
      \bottomrule
        \end{tabular}
        \end{center}
        \caption{\textbf{Question and Rationale Diversity.} Mean pairwise cosine distances in a sentence transformer space for various datasets. \ding{55} indicates lack of rationale. We choose one rationale per question on \env\ to make the comparison to other datasets with only one rationale. Rationales for VQAv2 come from the VQA-X. dataset~\cite{park2018multimodal}.}
        \label{table:BERTDist}
\end{table}

Finally, we use the same mean pairwise distance to look in particular at how different our questions are from OK-VQA which is the most similar prior work to ours. To do this we compare the minimum pairwise distance between every question in the OK-VQA training set to every question in the OK-VQA test set and our test set. We find that the average minimum distance from OK-VQA train to test is $\textbf{0.256}$ compared to $\textbf{0.311}$ between OK-VQA train and our test set\footnote{To make this comparison even, we chose a random subset of our test set to be the same size as OK-VQA test set so that the minimum is over the same number of possible choices in both cases.}. This shows that there is in fact a significant difference between our question set and OK-VQA in this feature space. 

\section{Experiments}

Next, we benchmark the \env\ dataset and compare the performance of different models. We consider three classes of methods: (1) \textbf{large-scale pre-trained models} such as CLIP~\cite{Radford2021LearningTV} and GPT-3~\cite{Brown2020LanguageMA}, (2) \textbf{models that generate and use rationales}, and (3) \textbf{specialized models} that are designed for knowledge-based VQA (KRISP~\cite{Marino2021KRISPII}) or tested for VQA (e.g., VilBERT~\cite{lu19vilbert}).

\subsection{Evaluation}
In the \emph{multiple choice (MC)} setting, a model chooses its answer from one of four options and we compute accuracy as the evaluation metric. In the \emph{direct answer (DA)} setting, a model can generate any text as its answer and we use the standard VQA evaluation from \cite{antol15}.

\subsection{Large-scale Pre-trained Models}\label{sec:large_models}

We compare three types of large-scale pre-trained models (discriminative, contrastive, and generative) in Table~\ref{tab:taskagn}. We also test these models in different input settings (where questions, images, or both are provided).

We compute BERT~\cite{Devlin2019BERTPO,hgf-bert} and CLIP ViT-B/32 text encoder representations for questions. We also compute ResNet-50~\cite{he2016deep} and CLIP ViT-B/32 features for images. These are provided as inputs to the appropriate discriminative and contrastive models. We provide questions as tokens and CLIP RN50x4 image representations as inputs to the generative models. We generate a vocabulary from a subset of training set answers and choices to use across all appropriate models. We describe this vocabulary further in Appx.~\ref{app:large}.

\textbf{Discriminative models.}
We train a multi-label linear classifier (i.e. MLP with one hidden layer and sigmoid activation function) on top of BERT (row \textit{d}), ResNet (row \textit{i}), and CLIP (rows \textit{e/j/m}) representations to score answers from the vocabulary. When questions and images are both provided, we first concatenate their representations. For the DA setting, we predict the top scoring vocabulary answer. For the MC setting, we instead predict the nearest neighbor\footnote{Cosine similarity between mean GloVe~\cite{pennington2014glove,hgf-glove} word embeddings.} choice to the top scoring vocabulary answer.

 \textbf{Contrastive models.}
We also evaluate models which match input questions and/or images with answers using their CLIP encodings. First, we evaluate the zero-shot setting (rows \textit{f/k/n}). If both questions and images are provided as inputs, we first add their representations. We select the answer whose encoding has the greatest cosine similarity to our input representation. We select from vocabulary answers in DA and the given choices in MC.

We also train a single-layer MLP on top of our input representations (rows \textit{g/l/o}). If both questions and images are provided, we first concatenate their representations. Our MLP produces a 512-d embedding and we train this with a CLIP-style contrastive loss between embeddings and their corresponding answers. We describe this loss further in Appx.~\ref{app:large}. We repeat the evaluation from the zero-shot setting, using these learned embeddings.

\setlength{\tabcolsep}{4pt}
% \begin{table}
% \begin{center}

% \begin{tabular}{lcccc}      
% \hline\noalign{\smallskip}

%  & \multicolumn{2}{c}{\textcolor{cl1}{Multiple Choice}} & \multicolumn{2}{c}{\textcolor{cl2}{Direct Answer}} \\
% \cmidrule(lr){2-3}\cmidrule(lr){4-5}
% \textbf{Method} & \textbf{Val} & \textbf{Test} & \textbf{Val} & \textbf{Test} \\

% \noalign{\smallskip}\hline\noalign{\smallskip}
\begin{table}[tp]
\begin{center}
\tabcolsep=0.2cm
\scriptsize
\begin{tabular}{l|cccc}      
\toprule
 & \multicolumn{2}{c}{\textcolor{cl1}{Multiple Choice}} & \multicolumn{2}{c}{\textcolor{cl2}{Direct Answer}} \\
\cmidrule(lr){2-3}\cmidrule(lr){4-5}
\textbf{Method} & \textbf{Val} & \textbf{Test} & \textbf{Val} & \textbf{Test} \\
\midrule
(a) Random & 26.70 & 25.36 & 0.03 & 0.06 \\
(b) Random (weighted) & 29.49 & \textbf{30.87} & 0.15 & 0.10 \\
(c) Most Common & \textbf{30.70} & 30.33 & \textbf{1.75} & \textbf{1.26} \\

\midrule
\textbf{Question} &  &  &  & \\

(d) BERT~\cite{Devlin2019BERTPO} (classifier) & 32.93 & 33.54 & 9.52 & 8.41 \\
(e) CLIP~\cite{Radford2021LearningTV} (classifier) & 32.74 & 33.54 & \textbf{13.10} & 10.24 \\
(f) CLIP~\cite{Radford2021LearningTV} (zero-shot) & 30.42 & 30.58 & 0.44 & 0.57 \\
(g) CLIP~\cite{Radford2021LearningTV} (contrastive) & \textbf{37.40} & \textbf{38.58} & 5.56 & 3.83 \\
(h) GPT-3~\cite{Brown2020LanguageMA} & 35.07 & 35.21 & 12.98 & \textbf{11.49} \\

%\noalign{\smallskip}
\midrule
\textbf{Image} &  &  &  & \\
(i) ResNet~\cite{he2016deep} (classifier) & 28.19 & 28.81 & 2.68 & 2.30 \\
(j) CLIP~\cite{Radford2021LearningTV} (classifier) & 33.21 & 32.56 & \textbf{5.15} & \textbf{4.38} \\
(k) CLIP~\cite{Radford2021LearningTV} (zero-shot) & \textbf{56.28} & \textbf{53.94} & 2.24 & 2.29 \\
(l) CLIP~\cite{Radford2021LearningTV} (contrastive) & 52.56 & 50.09 & 2.33 & 2.45 \\
%\noalign{\smallskip}
\midrule
\textbf{Question \& Image} &  &  &  &  \\
(m) CLIP (classifier) & 40.84 & 38.30 & 18.95 & 14.27 \\
(n) CLIP (zero-shot) & 48.19 & 45.72 & 1.08 & 0.71 \\
(o) CLIP (contrastive) & 53.77 & 51.01 & 10.36 & 7.10 \\
(p) ClipCap~\cite{Mokady2021ClipCapCP} & \textbf{56.93} & \textbf{51.43} & \textbf{30.89} & \textbf{25.90} \\
\bottomrule
%\hline

\end{tabular}
\end{center}
\caption{\textbf{Large-scale pre-trained models.} 
We also compare with no input heuristics (rows \textit{a-c}) with choices (for MC) or vocabulary answers (for DA). \textit{Random} is a uniform sampling. \textit{Random (weighted)} uses weighted sampling proportional to correct answer frequencies in train. \textit{Most Common} selects the most frequent answer in train.}
\label{tab:taskagn}
\end{table}

 \textbf{Generative models.}
We also evaluate models (GPT-3~\cite{brown2020language} and ClipCap~\cite{Mokady2021ClipCapCP}) that generate answers directly as text. For both models, we predict the generated text for DA and the generated text's nearest neighbor choice for MC.

We prompt GPT-3\footnote{We use the second largest available GPT-3 model, Curie, as in~\cite{west2021symbolic}.} (row \textit{h}) with 10 random questions and answers from the training set, followed by a new question, and let GPT-3 generate an answer to that question, in a manner similar to~\cite{west2021symbolic}.
We provide GPT-3 with the prompt template ``Question: ... Answer: [...]'', expecting it to complete the answer for each evaluation question.\footnote{During MC, we also tried prompting GPT-3 with ``Choices: ...'', but find that this actually hurts performance.}

ClipCap~\cite{Mokady2021ClipCapCP} (row \textit{p}) is an image captioning method that passes CLIP image features through a trained network to GPT-2 (as input tokens). We adapt this model by adding question tokens (and answer choices if applicable) to the prompt of GPT-2, generate answers instead of captions, and fine-tune on our data. We provide additional details, diagrams, and variations in Appx.~\ref{app:large}.

 \textbf{Results.}
Table~\ref{tab:taskagn} shows the results of our evaluation of these models. Rows \textit{a-c} show the biases in our dataset, but that the direct answer setting is appropriately challenging. Question-only baselines (rows \textit{d-h}) show poor performance in both MC and DA settings. However, it is interesting that GPT-3 performs similarly to the fine-tuned CLIP models (whichever is better per setting). The zero-shot CLIP model (row \textit{f}) is least effective, indicating that training is necessary to repurpose CLIP text encodings for language-only tasks.
Unsurprisingly, CLIP image features are very strong for zero-shot multiple choice matching (row \textit{k}). However, they are not as strong as for the fine-tuned classifier (row \textit{j}) in DA. ClipCap (row \textit{p}) outperforms all other baselines in DA, because we use powerful image features and also fine-tune a strong language model for our task.

\subsection{Rationale Generation}\label{sec:rationale_gen}
\setlength{\tabcolsep}{4pt}
% \begin{table}
% \begin{center}
% \caption{}
% \label{tab:rationale_baselines}

% \begin{tabular}{lcccc}      
% \hline\noalign{\smallskip}

%  & \multicolumn{2}{c}{\textcolor{cl1}{Multiple Choice}} & \multicolumn{2}{c}{\textcolor{cl2}{Direct Answer}} \\
% \cmidrule(lr){2-3}\cmidrule(lr){4-5}
% \textbf{Method} & \textbf{Val} & \textbf{Test} & \textbf{Val} & \textbf{Test} \\

% \noalign{\smallskip}\hline\noalign{\smallskip}

\begin{table}[tp]
\begin{center}
\tabcolsep=0.2cm
\scriptsize
\begin{tabular}{l|cccc}      
\toprule
& \multicolumn{2}{c}{\textcolor{cl1}{Multiple Choice}} & \multicolumn{2}{c}{\textcolor{cl2}{Direct Answer}} \\
\cmidrule(lr){2-3}\cmidrule(lr){4-5}
\textbf{Method} & \textbf{Val} & \textbf{Test} & \textbf{Val} & \textbf{Test} \\
\midrule
(a) ClipCap $\rightarrow$ Cap. $\rightarrow$ GPT & 42.51 & 43.61 & 16.59 & 15.79 \\
(b) ClipCap $\rightarrow$ Ratl. $\rightarrow$ GPT & \textbf{44.00} & \textbf{43.84} & \textbf{18.11} & \textbf{15.81} \\
\midrule
\textbf{Oracles} & & & & \\
(c) GT Caption $\rightarrow$ GPT & 45.40 & --- & 16.39 & --- \\
(d) GT Rationale $\rightarrow$ GPT & \textbf{56.74} & 56.75 & \textbf{24.02} & 20.75 \\

\bottomrule
\end{tabular}
\end{center}
\caption{\textbf{Models using generated and GT rationales} as described in Sec.~\ref{sec:rationale_gen}. We are unable to evaluate the GT Caption $\rightarrow$ GPT setting on the test set, as captions are not available in the COCO~\cite{Chen2015MicrosoftCC} test set.}
\label{tab:taskagn_rationales}
\end{table}

We are interested in whether we can improve GPT-3 prompting results by providing additional image- and question- specific context and report results for the following methods in Table~\ref{tab:taskagn_rationales}. So, we fine-tune ClipCap (given images and questions, but not choices) as above, but for the task of generating rationales instead of answers. Our model scores $\textbf{10.2}$ (val) / $\textbf{9.58}$ (test) on SacreBLEU~\cite{post-2018-call} and $\textbf{0.271}$ (val) / $\textbf{0.256}$ (test) on METEOR~\cite{banerjee-lavie-2005-meteor}. We can then prompt GPT-3 (as above) but also provide these generated rationales as ``Context: ...''. This model is denoted by `ClipCap $\rightarrow$ Ratl. $\rightarrow$ GPT'. We provide additional details, diagrams, and examples of generated rationales in Appx.~\ref{app:rationale}. We repeat this experiment using captions (generated from only images) from the original ClipCap model: `ClipCap $\rightarrow$ Cap. $\rightarrow$ GPT'.

 \textbf{Results.} We show results from these experiments in Table~\ref{tab:taskagn_rationales}. Interestingly, prompting GPT-3 with ground-truth rationales (row \textit{d}) is competitive with the best model in Sec.~\ref{sec:large_models} (Table~\ref{tab:taskagn}, row \textit{p}) in MC and significantly outperforms the question-only GPT-3 method (Table~\ref{tab:taskagn}, row \textit{h}). When we prompt GPT-3 with ground-truth rationales (row \textit{d}), we see higher performance than when we provide ground-truth captions (row \textit{c}). This affirms that rationales contain useful information (i.e. specific to our questions and answers) in addition to captions. However, the additional performance of prompting GPT-3 using generated rationales (row \textit{b}) over generated captions (row \textit{a}) is not as significant. This indicates potential room for improvement in our approach for generating rationales.

\subsection{Specialized Models}\label{sec:specialized_models}

\begin{table}[tp]
\begin{center}
\tabcolsep=0.2cm
\scriptsize
\begin{tabular}{l|cccc}      
\toprule
 & \multicolumn{2}{c}{\textcolor{cl1}{Multiple-Choice}} & \multicolumn{2}{c}{\textcolor{cl2}{Direct Answer}} \\
\cmidrule(lr){2-3}\cmidrule(lr){4-5}
\textbf{Method} & \textbf{Val} & \textbf{Test} & \textbf{Val} & \textbf{Test} \\
\midrule
(a) Pythia~\cite{Jiang2018PythiaVT} &  49.0 & 40.1  & 25.2 & 21.9 \\ 
(b) ViLBERT~\cite{lu19vilbert} - OK-VQA &  32.8 & 34.1  & 9.1 & 9.2 \\ 
(c) ViLBERT~\cite{lu19vilbert} - VQA&  47.7 & 42.1 &17.7 & 12.0 \\ 
(d) ViLBERT~\cite{lu19vilbert} &  49.1 & 41.5  & 30.6 & 25.9 \\
(e) LXMERT~\cite{Tan2019LXMERTLC} &  51.4 & 41.6 & 30.7 & 25.9 \\ 
(f) KRISP~\cite{Marino2021KRISPII} &  51.9 & 42.2 & 33.7 & 27.1 \\ 
(g) GPV-2~\cite{Kamath2022WeblySC} & \textbf{60.3}  & \textbf{53.7} & \textbf{48.6} & \textbf{40.7} \\ 
%GPV-2~\cite{Kamath2022WeblySC} + Rat. & 70.5 & 44.0 & 66.4 & 36.4 \\ 
\midrule
\textbf{Oracles} & & & & \\
(h) GPV-2~\cite{Kamath2022WeblySC} + Masked Ans. & 65.1  & 58.3 & 52.7 & 43.9 \\ 
(i) GPV-2~\cite{Kamath2022WeblySC} + GT Ratl. & 73.4 & 67.2 & 58.9 & 51.7 \\ 

\bottomrule
\end{tabular}
\end{center}
\caption{\textbf{Specialized models results. } Baselines trained for VQA or knowledge-based VQA, and fine-tuned on \env. The bottom two rows are not comparable with the others since they use ground-truth rationales at test time. }
\label{tab:taskspc}

\end{table}

In this section, we evaluate some recent high-performing, open-source models trained on knowledge-based VQA or the traditional VQA. The models we consider are Pythia~\cite{Jiang2018PythiaVT}, VilBERT~\cite{lu19vilbert}, LXMERT~\cite{Tan2019LXMERTLC}, KRISP~\cite{Marino2021KRISPII}, and GPV-2~\cite{Kamath2022WeblySC}. As the first four models are part of MMF~\cite{singh2020mmf}, it is easier to compare them fairly. KRISP is a high-performing model on OK-VQA~\cite{marino19cvpr}. It provides a suitable baseline as it addresses knowledge-based VQA. GPV-2 performs multiple vision and vision--language tasks and has learned a large number of concepts, so it can be a strong baseline for \env. All of these models are fine-tuned on \env\ to predict answers directly for DA evaluation. We adapt them to MC using the nearest choice method described above. See Appx.~\ref{app:special} for the details of each model.

\textbf{Results.} Unsurprisingly, these models, which are specialized for DA and some of which are specialized for knowledge-based VQA perform very well on the DA evaluation and quite well on MC. Of the models trained only on \env\, KRISP does the best, likely because it is trained to directly use outside knowledge graphs. GPV-2, however, performs best of all, beating all other models (that do not use ground-truth rationales) in all settings, possibly because of the large number of concepts it has learned.

\textbf{Transfer results.}
We train ViLBERT on VQAv2 and OK-VQA datasets (denoted by `ViLBERT-VQA' and `ViLBERT-OK-VQA' in Table~\ref{tab:taskspc}) to evaluate whether the knowledge from those datasets is sufficient for \env. The low performance shows the significant difference between these datasets.

\textbf{Ground-truth Rationales.}
To evaluate how well the model performs if it is provided with high-quality rationales, we use ground-truth rationales at test. We show these results with GPV-2 (our best model). Ground-truth rationales are appended to questions as additional input text (`GPV-2 + GT Ratl.'). For this experiment, we used only one of the rationales. Comparing rows \textit{g} and \textit{i} of Table~\ref{tab:taskspc} shows rationales are helpful. To evaluate how much of this improvement can be attributed to rationales and not the fact that sometimes rationales contain the answer, we replaced answers in the rationales with $[$answer$]$ token. The performance drops (row \textit{i} vs row \textit{h}), however, it is still higher than the case that we do not use rationales (row \textit{h} vs row \textit{g}).

\section{Analysis of Models}
Next, we analyze the predictions that our baseline models make to see if we can learn more about \env: what kinds of questions do different types of approaches do better / worse on? For these experiments, we choose some of the best performing models on Direct Answer: VilBERT~\cite{lu19vilbert}, LXMERT~\cite{Tan2019LXMERTLC}, KRISP~\cite{Marino2021KRISPII},  ClipCap~\cite{Mokady2021ClipCapCP} and GPV-2~\cite{Kamath2022WeblySC}. We also use the ClipCap $\rightarrow$ Rationale $\rightarrow$ GPT model from Table~\ref{tab:taskagn_rationales}, which will be referred to as `GR-GPT' for Generated Rationales GPT.

\textbf{Answer Frequency.}
First, we look at how answer frequency affects performance in Table~\ref{table:AnsFreqAcc}. We first count the number of times any answer appears in the direct answers in the training set. We then divide these into bins and look at the direct DA test accuracy of our baselines for each of these frequency bins. We find that GPV-2, and to a lesser extent ClipCap and GR-GPT perform better on questions whose answers do not appear often in the training set (1-5 and 6-10 columns of Table~\ref{table:AnsFreqAcc}). GPV-2 in particular (which is fine-tuned on several vision and language tasks) is able to predict these tail answers much better than other methods, especially the discriminative methods such as LXMERT.

\begin{table}[tp]
\scriptsize
 \setlength{\tabcolsep}{5pt}
     \begin{center}
        \begin{tabular}{l | c c c c c c c}
        \toprule
      Model & 1-5 & 6-10 & 11-20 & 21-50 & 51-100 & 101-200 & 201+\\ \midrule 
      VilBERT~\cite{lu19vilbert} & 0.00 & 0.00 & 3.68 & 10.97 & 19.95 & 26.53 & 35.91 \\
      LXMERT~\cite{Tan2019LXMERTLC} & 0.00 & 0.00 & 4.29 & 13.73 & 20.18 & 26.69 & 34.31 \\
      KRISP~\cite{Marino2021KRISPII} & 0.00 & 0.61 & 6.34 & 13.99 & 21.78 & 28.55 & 35.22 \\
      ClipCap~\cite{Mokady2021ClipCapCP} & 4.71 & 4.24 & 9.10 & 17.90 & 25.93 & 29.44 & 33.99 \\
      GR-GPT & 8.18 & 9.29 & 9.41 & 17.39 & 18.31 & 21.98 & 24.65\\
      GPV-2~\cite{Kamath2022WeblySC} & \textbf{10.16} & \textbf{12.12} & \textbf{22.60} & \textbf{31.04} & \textbf{38.40} & \textbf{41.60} & \textbf{44.69} \\
      \bottomrule
        \end{tabular}
        \end{center}
        \caption{\textbf{Results across different answer frequencies.} The questions are categorized based on the frequency of the GT answer in the training set. Columns show accuracy for answers that appear 1-5 times, 6-10 times, etc. If multiple direct choices, we default to most common one.}
        \label{table:AnsFreqAcc}

\end{table}

\textbf{Knowledge Types.}
Next, we use the subset of test that we collected knowledge types on (see Sec.~\ref{sec:dataset_stats}) to look at the direct answer accuracy of these models for different types of knowledge. In Table~\ref{table:KnowledgeTypeAcc}, we see that while again GPV is the best overall and in every category, the results show some interesting distinctions. KRISP, which is specifically designed with access to explicit knowledge sources such as ConceptNet~\cite{liu2004conceptnet} performs better on ``Knowledge Base'' questions compared with other discriminative multi-modal transformer methods such as VilBERT and LXMERT as well compared to ClipCap which has an overall higher performance. It also performs better on ``Physical Knowledge'' which also tends to overlap with the knowledge sources it has.

\newcommand{\smalltablefont}[1]{\fontsize{7}{12}\selectfont{#1}}
\begin{table}[tp]
\scriptsize
 \setlength{\tabcolsep}{5pt}
     \begin{center}
        \begin{tabular}{l | c c c c}
        \toprule

      Model & Commonsense & Knowledge Base & Physical Knowledge & Visual Knowledge\\ \midrule 
      VilBERT~\cite{lu19vilbert} & 24.30 & 19.96 & 29.76 & 26.55 \\
      LXMERT~\cite{Tan2019LXMERTLC} & 25.51 & 16.01 & 27.38 & 27.23 \\
      KRISP~\cite{Marino2021KRISPII} & 26.63 & 20.72 & 39.29 & 26.09 \\
      ClipCap~\cite{Mokady2021ClipCapCP} & 27.19 & 16.57 & 30.95 & 33.41 \\
      GR-GPT & 21.42 & 12.99 & 17.86 & 24.79 \\
      GPV-2~\cite{Kamath2022WeblySC} & 39.76 & 25.24 & 44.05 & 41.19\\
      \bottomrule
        \end{tabular}
        \end{center}
        \caption{\textbf{Analysis of results based on knowledge type.} }
        \label{table:KnowledgeTypeAcc}

\end{table}

\textbf{Prediction overlap/difference.} Finally, we look at some statistics on a question by question level in the \env\ test set. Specifically we look at the overlap in which methods answered which questions correctly\footnote{For ease of analysis we count a binary yes/no of whether a model answered correctly if it answered any possible answer in the direct answer set.}. We use the same models as in Tables~\ref{table:AnsFreqAcc} \& \ref{table:KnowledgeTypeAcc}.

First, we find that only $\textbf{5.85\%}$ of questions in test were answered correctly by all models and $\textbf{30.96\%}$ of questions had no model predict a correct answer for. Considering the worst performing model of these gets $\textbf{15.81\%}$ DA accuracy and the best gets $\textbf{40.7\%}$, it implies that there is actually a large variation between these models beyond some just being generally better than others and thus getting ``hard'' questions right and keeping performance on ``easy'' questions. 

In Table~\ref{table:PairwiseAnsDiff}, we show the difference between the questions each model gets right on \env\ test. Each row shows the percentage of that method's correctly answered questions that were not correctly answered by the comparison model in each column. If we look at the row for the lowest performing model (GR-GPT) for the column for the best performing model (GPV-2), we still see that $\textbf{29.2\%}$ of GR-GPT's correctly answered questions are answered wrong by GPV-2!

Finally, to further illustrate the point that different models have very different mistake patterns, we take the prediction of all of these models except for GPV-2 for each question and take the majority vote between these. This majority vote combination gets an accuracy of $\textbf{29.5}$ compared to the best of these models which gets $\textbf{27.1}$. This does not work when GPV-2 is added (this majority model gets $\textbf{35.60}$ which is lower than GPV-2's $\textbf{40.7}$). We can also look at the Oracle combination accuracy. That is, from our six models, choose the answer with the highest ground-truth value and take that as the oracle combination answer. This DA accuracy is $\textbf{56.87}$ versus the single best performance of $\textbf{40.7}$, again showing that even worse performing models get lots of questions right that the best model gets wrong.

\begin{table}[tp]
\scriptsize
 \setlength{\tabcolsep}{5pt}
     \begin{center}
        \begin{tabular}{l | c c c c c c}
        \toprule
      Model & VilBERT & LXMERT & KRISP & ClipCap & GR-GPT & GPV-2 \\ \midrule 
      VilBERT~\cite{lu19vilbert} & 0.00 & 29.00 & 27.19 & 43.72 & 59.72 & 26.33 \\
      LXMERT~\cite{Tan2019LXMERTLC} & 28.07 & 0.00 & 26.57 & 44.39 & 59.73 & 27.44 \\
      KRISP~\cite{Marino2021KRISPII} & 30.44 & 30.76 & 0.00 & 44.18 & 60.29 & 27.43 \\
      ClipCap~\cite{Mokady2021ClipCapCP} & 48.72 & 49.98 & 46.76 & 0.00 & 55.94 & 26.64 \\
      GR-GPT & 50.27 & 50.91 & 48.67 & 40.30 & 0.00 & 29.20 \\
      GPV-2~\cite{Kamath2022WeblySC} & 51.09 & 52.46 & 49.57 & 46.56 & 61.94 & 0.00 \\
      \bottomrule
        \end{tabular}
        \end{center}
        \caption{\textbf{Pairwise difference between correctly answered questions.} For row $i$ and column $j$ of this table the value is percentage of questions answered correctly by model $i$ that $j$ did not answer correctly. }
        \label{table:PairwiseAnsDiff}

\end{table}

\begin{figure}[tp]
    \centering
    \includegraphics[width=33pc]{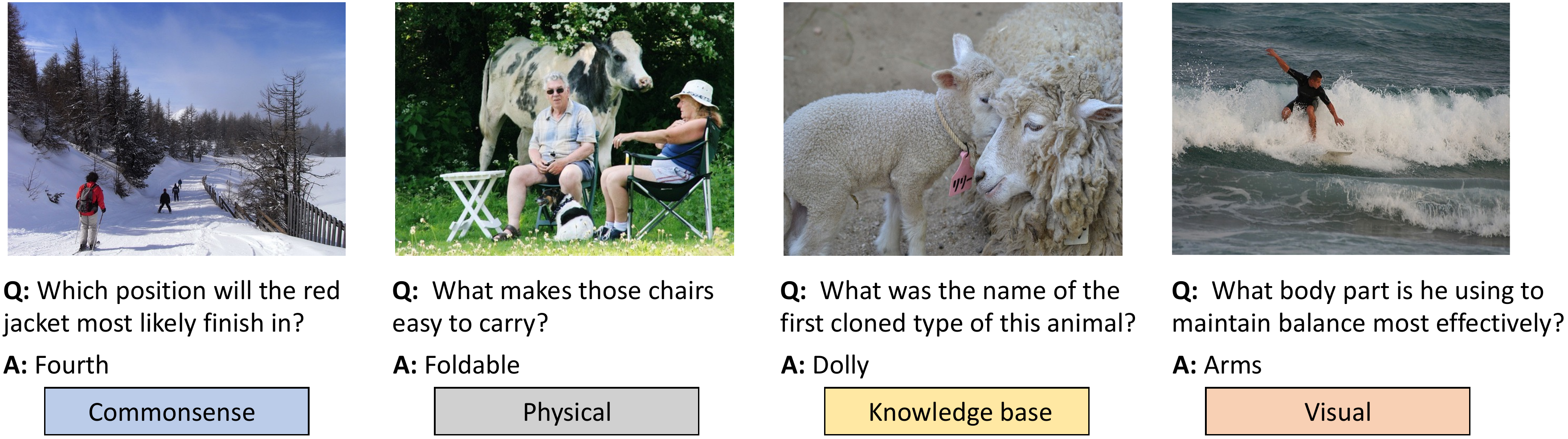}
    \caption{\textbf{Example questions that all discussed models fail at.} }
    \label{figures:nomodel} 
\end{figure}

\textbf{Qualitative Analysis.} We analyzed our models and extracted questions that all of the discussed models fail at. Figure~\ref{figures:nomodel} shows an example from each knowledge type. This qualitative example shows what type of reasoning is missing in our current top performing models. 

All of these analyses together provide several interesting findings. First, aside from being generally difficult, the \env\ dataset shows a surprising lack of overlap in the specific questions different models answer correctly. Second, we see that different methods handle rare answers very differently. Moreover, different methods perform differently based on the type of knowledge. All of this suggests that \env\ provides many different kinds of challenging questions which bring out different strengths and weaknesses of methods.

\section{Conclusion}

Vision and language models have become progressively more powerful, however, evaluation of the reasoning capabilities of these models have not received adequate attention. To take a step in this direction, we propose a new knowledge-based VQA benchmark called \env, which primarily includes questions that require reasoning using commonsense and world knowledge. We provide \textit{rationales} for each question so models can learn the line of reasoning that leads to the answer. We evaluate a large set of recent, high performance baselines. While they show impressive performance on the proposed task, it is evident that they lack the reasoning capability and/or the knowledge required to answer the questions, and there is a large room for improvement. Through extensive analyses, we show different models have different weaknesses and strengths. To solve \env \ and to move towards general multi-modal intelligence, we need to combine many types of capabilities from many different methods.

{\small
\bibliographystyle{ieee}
\bibliography{egbib}
}

\appendix
\section{Additional details of dataset collection}

\subsection{Examples of rejected questions}\label{sec:supp_rej_questions}

\begin{figure}[th]
    \centering
    \includegraphics[width=\linewidth]{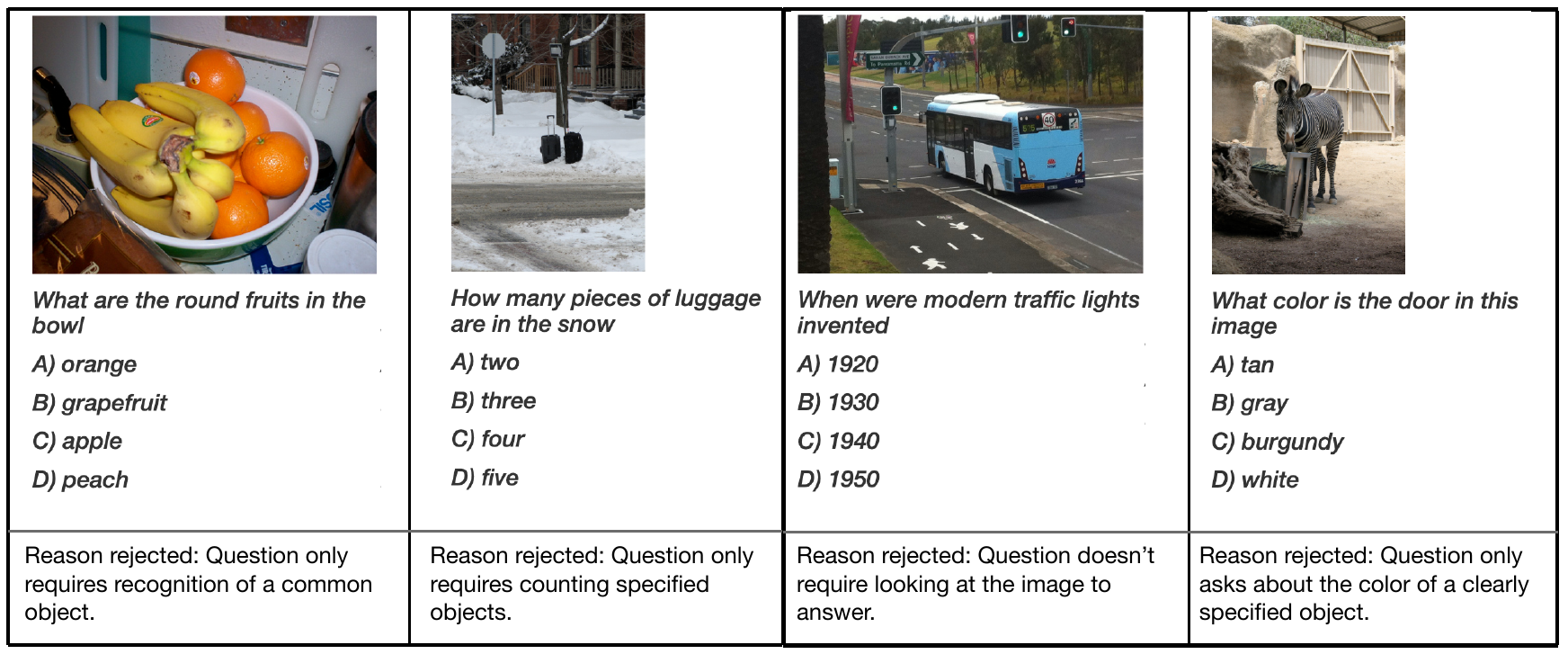}
    \caption{Examples of questions rejected for not meeting our criteria.}
    \label{fig:rej_q}
\end{figure}

With a focus on overall question quality, we removed around 60\% of questions written for having any of several flaws. The vast majority of questions removed exhibited one or more four flaws: 1) Only required recognition of a common object, 2) only required counting a readily specified object, 3) did not require looking at the image to answer, 4) only asked about the color of a readily specified object. Examples of questions from each of these categories are shown in Fig.~\ref{fig:rej_q}.

\subsection{Data collection interface}\label{sec:supp_rej_questions}

\begin{figure}[th]
    \centering
    \includegraphics[width=\linewidth]{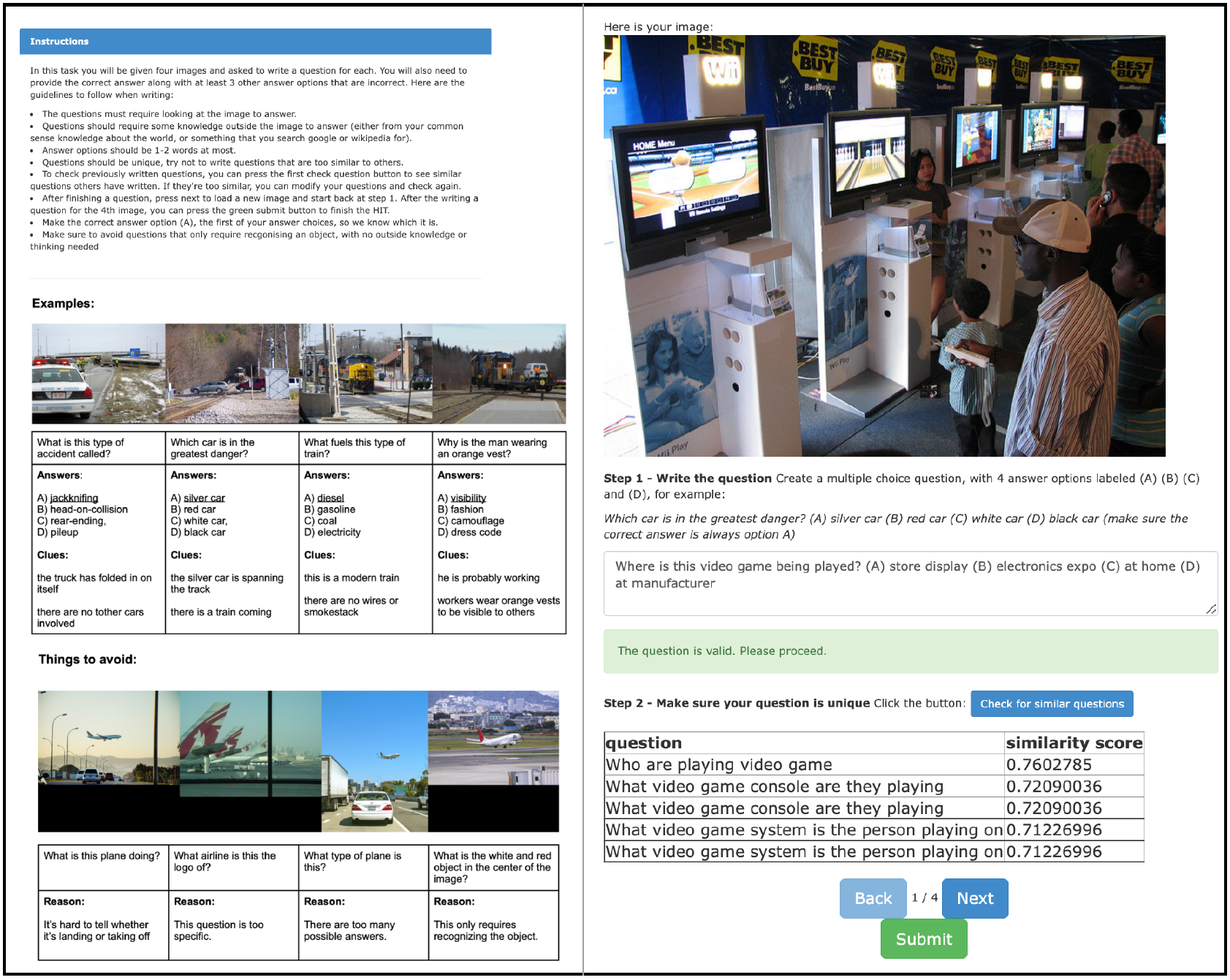}
    \caption{Instructions and interface used for question collection.}
    \label{fig:q_int}
\end{figure}

The data-collection interface used by crowdworkers to write questions is shown in Fig.~\ref{fig:q_int}. Detailed instructions along with examples of good and bad questions were provided. After writing a question, workers were required to press the ``Check for similar question" button. This sent a request to a server which returned the five questions closest to those already written in our growing dataset. We asked workers to rewrite or rephrase questions that were too similar, but did not enforce a minimum distance cutoff. The set of questions queried were reset when collecting the val and test sets to allow a greater degree of overlap with the training set. After satisfied with their question, workers advanced to the next image. Each task workers performed included four images, nearby neighbors in a CLIP embedding space, which encouraged creative differences in questions written for similar images. Workers were only required to write two questions (out of four possible images) to allow them to skip images they didn't feel they could write a suitable questions for. This cut down on unsuitable questions that they would have otherwise been forced to write in order to complete the task. After completing two questions, workers were allowed to submit their work and advance to the next image set.

\begin{figure}[th]
    \centering
    \includegraphics[width=\linewidth]{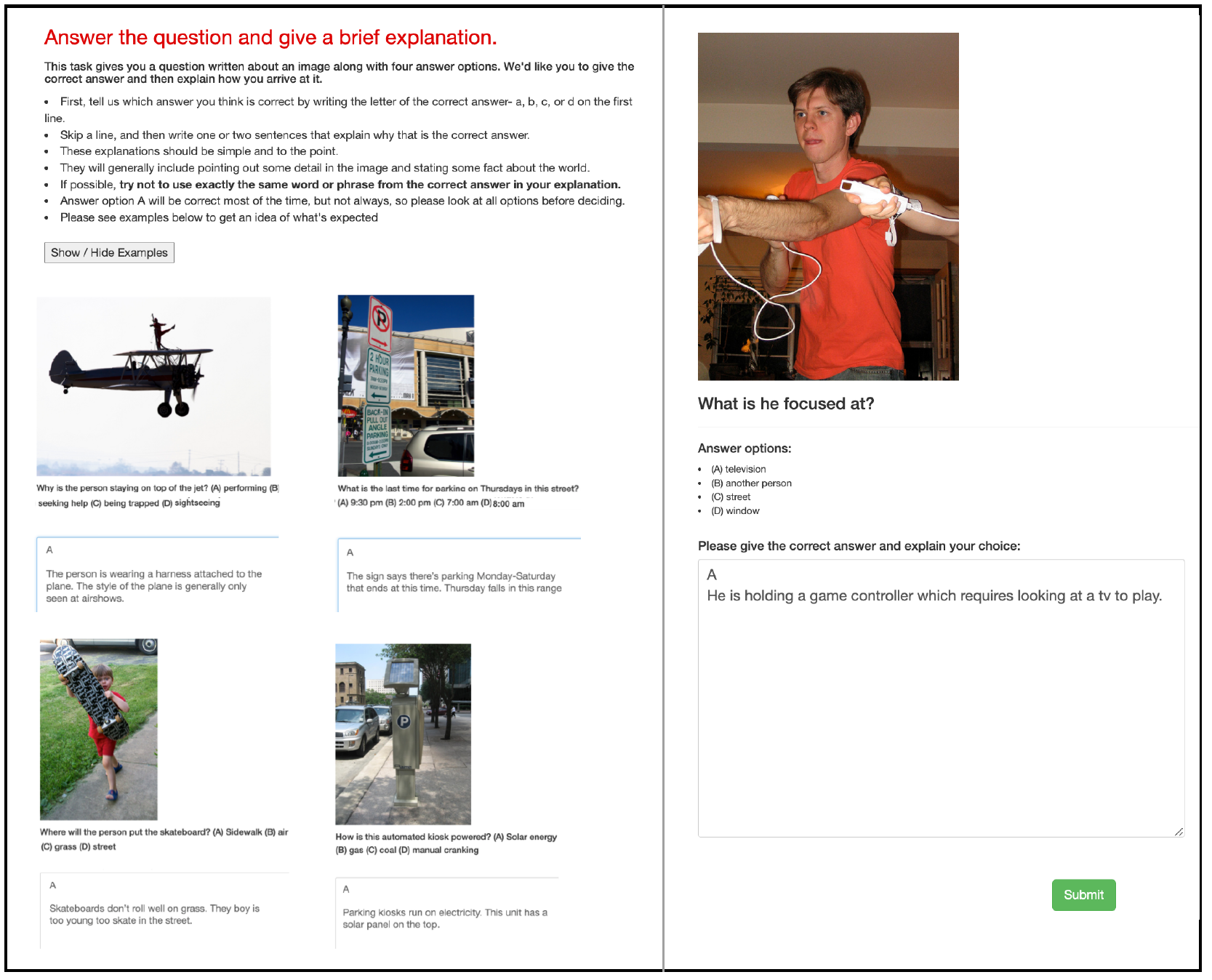}
    \caption{Instructions and interface used for rationale collection.}
    \label{fig:r_int}
\end{figure}

The data-collection interface used by crowdworkers to write rationales is shown in Fig.~\ref{fig:r_int}. Detailed instructions along with examples of good rationales were provided. We first asked workers to confirm the correct answer or provide the answer they thought was correct. This allowed a check on the correctness of the original question, and questions with a disagreement were removed from the dataset. Workers then provided a 1-2 sentence explanation of why the answer was correct that included any external knowledge needed to arrive there.

\section{Additional Details for Large-scale Pre-trained Models}

We produce the vocabulary for the experiments in Sec.~\ref{sec:large_models} from the training set by selecting all correct choices, as well as all choices and direct answers that appear in at least three questions. This results in a vocabulary with 10,424 answers.

\label{app:large}
\subsection{Discriminative models}

We train all of our discriminative models for 500 epochs with a learning rate of $0.01$ and batch size of $128$, except the model with ResNet input features, which is trained with a learning rate of $0.001$.

\subsection{Contrastive models}\label{sec:supp_contrastive}

\begin{figure}[th]
    \centering
    \includegraphics[width=\linewidth]{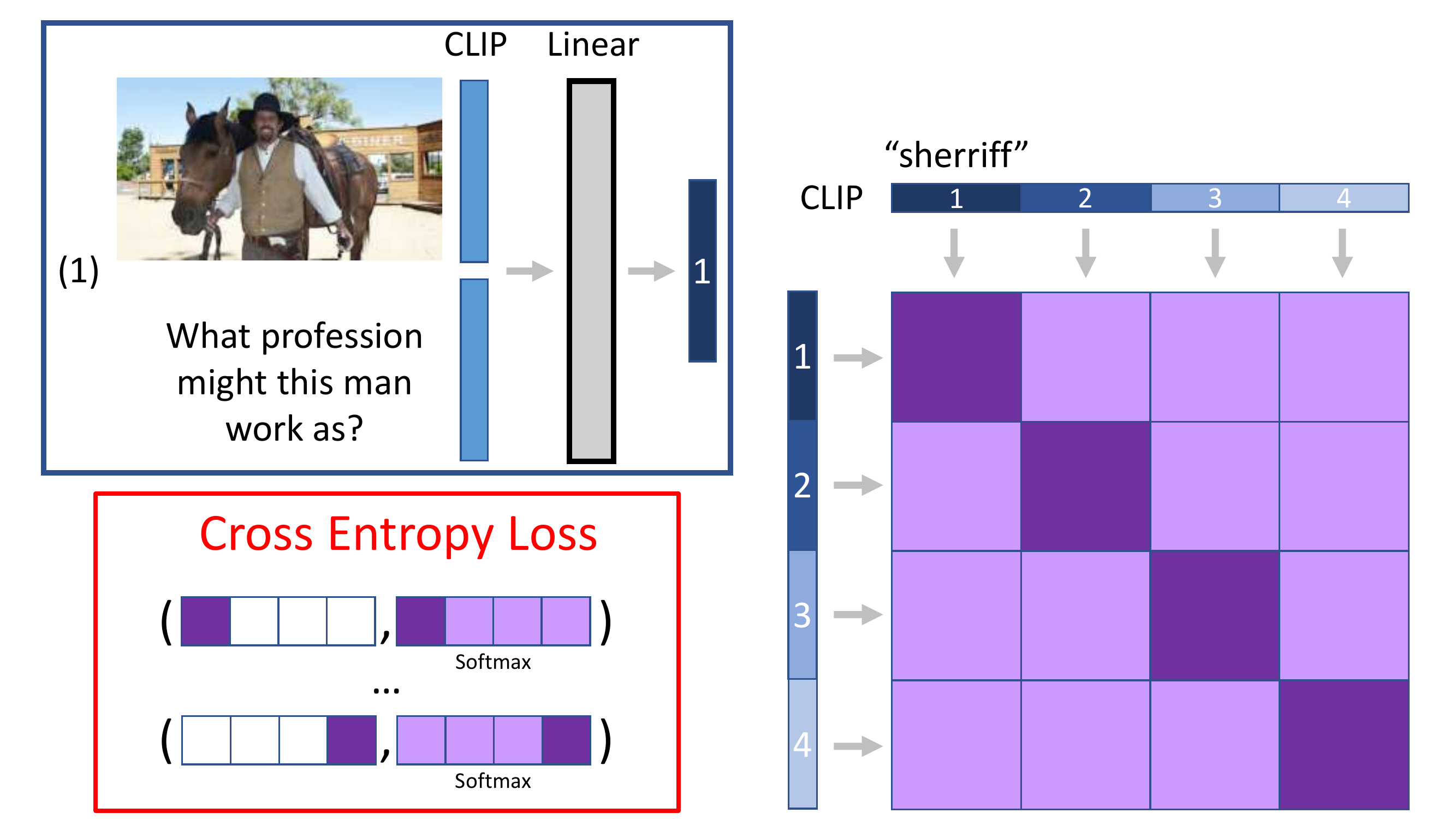}
    \caption{As described in Sec.~\ref{sec:supp_contrastive}. CLIP-style contrastive loss between embeddings (of questions and images) and CLIP text encodings (of answers). Shown for a batch size of 4.}
    \label{fig:contrastive}
\end{figure}

The CLIP zero-shot setting requires no training. In the trained setting, we train our linear layer for 500 epochs with a learning rate of $0.01$ and batch size of $128$. We further elaborate on our ``CLIP-style contrastive loss'' below and visualize it in Fig.~\ref{fig:contrastive}.

Recall that we have passed CLIP representations (for questions and/or images) through a linear layer to produce a 512-d embedding (the same size as a CLIP text encoding). For a batch of embeddings $E$ and the CLIP text encodings of their corresponding answers $A$, we produce a cosine similarity matrix between $E$ and $A$ (i.e. the purple matrix in Fig.~\ref{fig:contrastive}, showing a batch size of 4). We apply softmax over each matrix row (producing embedding--answer matching probabilities per embedding over answers in $A$) and compute a cross-entropy loss to maximize the similarity between each embedding and its corresponding answer.

\subsection{Generative models}\label{sec:supp_generative}

\begin{figure}[th]
    \centering
    \includegraphics[width=\linewidth]{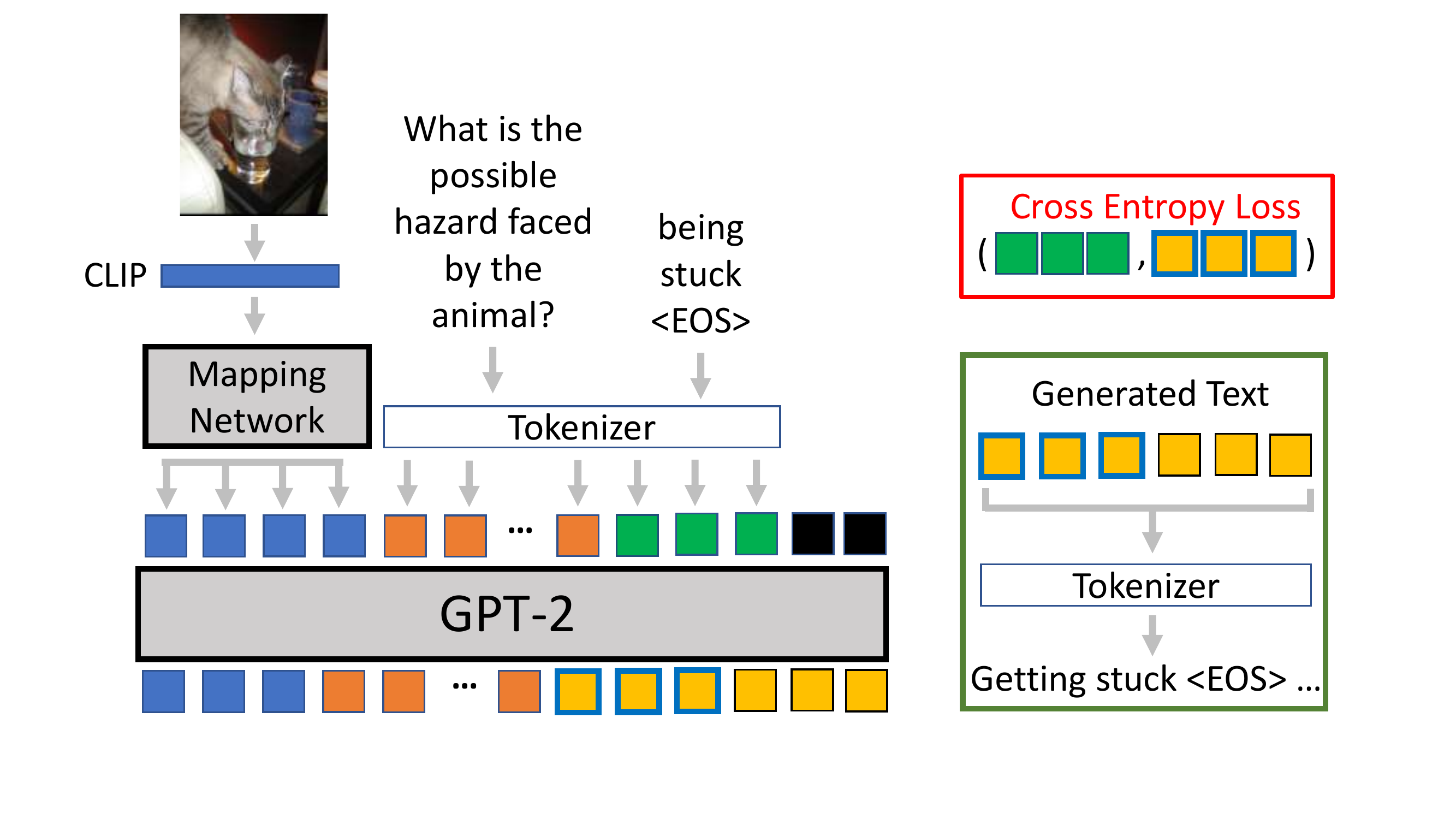}
    \caption{Diagram of modified ClipCap architecture for VQA tasks.}
    \label{fig:clipcap}
\end{figure}

We show our modified ClipCap model in Fig.~\ref{fig:clipcap}. As in ClipCap~\cite{Mokady2021ClipCapCP}, we provide CLIP image representations to a mapping network, which produces prefix tokens as input for GPT-2. We then tokenize our question and ground-truth answer (appended with an end-of-sequence string, $\langle\textrm{EOS}\rangle$) and also provide these tokens as input. The remaining input tokens (in black) are zero-padding. As mentioned in our paper, we also appended the (pre-tokenized) question string with ``Choices: ...'' during the MC setting.

This model is trained autoregressively. I.e., $O_i$ is generated conditionally, given $I_0 \cdots I_i$ (for input tokens $I$ and output logits $O$), and supervised with a cross-entropy loss against the next sequence token $I_{i+1}$. In our case, we only compute this cross-entropy loss for outputs corresponding with the ground-truth answer tokens (including $\langle\textrm{EOS}\rangle$).

At inference time, we prompt GPT-2 with our image prefix and question tokens. We have the model predict the most likely next token (i.e. generating a token in the answer) from the output logits. We append this token to the input and repeat this step, until the model predicts $\langle\textrm{EOS}\rangle$. We can use the tokenizer to decode these output tokens (excluding $\langle\textrm{EOS}\rangle$), producing our model's textual answer prediction. Note that beam search is an alternative way to generate text from autoregressive language models, but we found that it led to worse results, likely because the answers we are trying to generate are short (e.g. 1-3 words).

We fine-tuned the models in our experiments (choosing the checkpoint with the best F1 validation score for generated answers over 10 epochs), using the settings and COCO pre-trained weights (for the MLP mapping network) made available by the ClipCap authors~\footnote{\url{https://github.com/rmokady/CLIP_prefix_caption}}. For the pre-trained MLP model, they used CLIP ViT-B/32 features, produced 10 image prefix tokens, and had also fine-tuned GPT-2 (for their image captioning task). We further fine-tuned the GPT-2 weights on our task.

\section{Additional Details for Rationale Generation}
\label{app:rationale}
We generated rationales from ClipCap in a nearly identical manner to how we generated answers (see Sec.~\ref{sec:supp_generative} and Fig.~\ref{fig:clipcap} above). However, we replace the ground-truth answer string/tokens with a ground-truth rationale. And, we don't provide ``Choices: ...'' in the ClipCap prompt for the MC setting. We also use beam search during generation, as it seems to perform better for these longer strings. We also use the MLP mapping network and continue to fine-tune GPT-2, as it demonstrates the best performance for this task. We again fine-tuned this model on our training data for 10 epochs and picked the checkpoints with best BLEU and METEOR validation scores.

\begin{figure}[th]
    \centering
    \includegraphics[width=\linewidth]{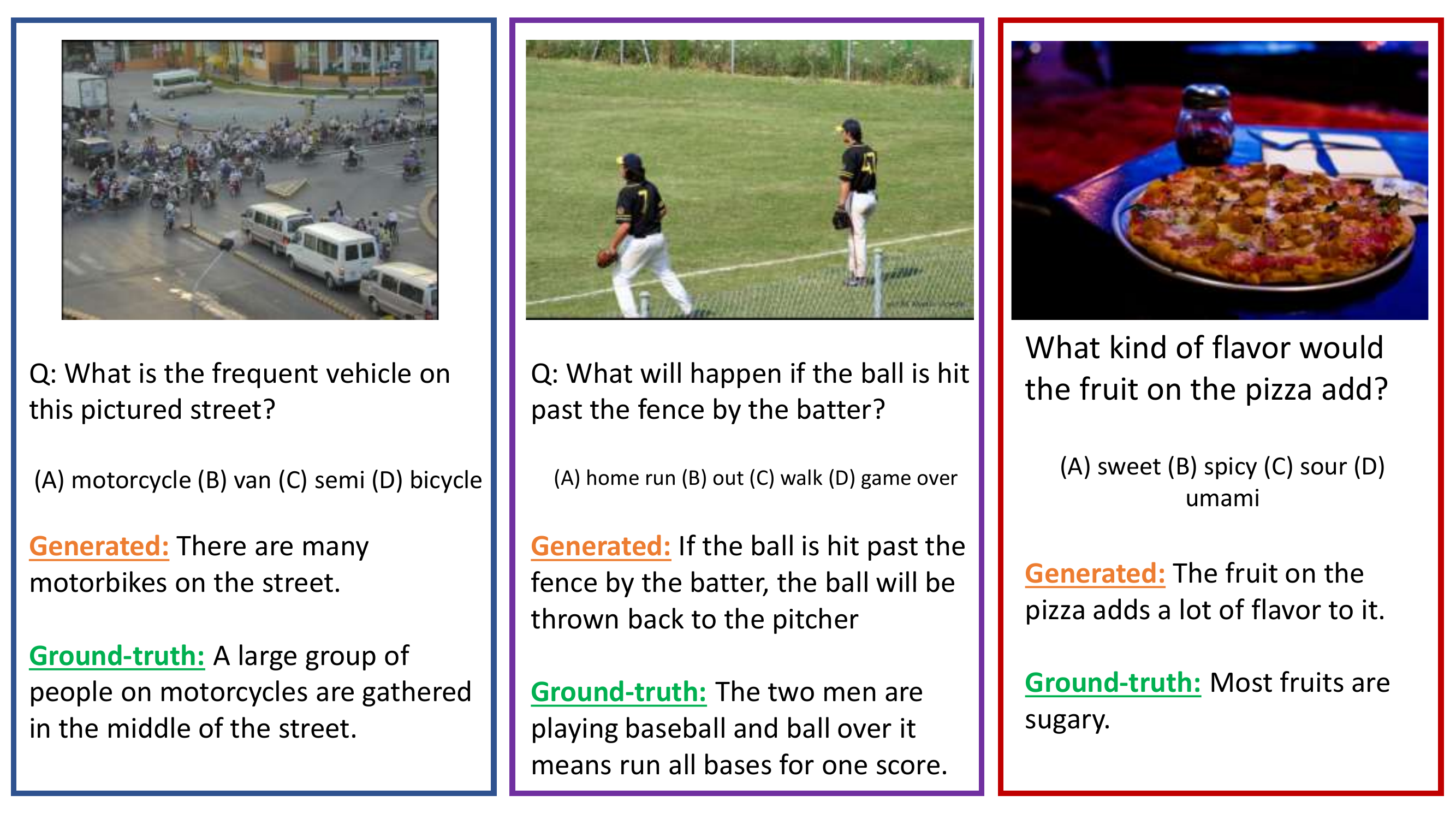}
    \caption{Examples of rationales generated by our modified ClipCap method for examples in our validation set.}
    \label{fig:rationale_examples}
\end{figure}

We show some examples of generated rationales in Fig.~\ref{fig:rationale_examples}.

\section{Additional Details for Specialized Models}
\label{app:special}
For all of these models, we use the same training hyperparameters as the original implementation. For all of the discriminative methods in the paper we use a fixed vocabulary constructed from direct answers that appeared two or more times in the training set. This includes  2,133 bi-grams or unigrams, with 1,937 words.

\textbf{Pythia~\cite{Jiang2018PythiaVT}} Pythia is a modification of \cite{anderson2018bottom} that introduces changes to the architecture and learning schedule and utilizes more training data. We fine-tune it on the \env\ dataset. For fine-tuning, we replace the top classification layer with a randomly initialized layer for our set of answer vocabulary.  

\textbf{LXMERT~\cite{Tan2019LXMERTLC}} LXMERT is a Transformer-based vision and language model pre-trained using a large amount of image-sentence pairs for a set of pre-training tasks such as masked language modeling and object prediction. The model is pre-trained on VQAv2~\cite{goyal2017making}, GQA~\cite{Hudson2019GQAAN}, VG-QA~\cite{zhu16}, COCO captions~\cite{Chen2015MicrosoftCC}, and Visual Genome captions~\cite{krishna17}. We then fine-tune the model using the training set of \env. 

\textbf{VilBERT~\cite{lu19vilbert}} ViLBERT is an extension of the BERT architecture to process vision and language modalities for learning a joint representation for them. ViLBERT has been pre-trained on proxy tasks, but it has been evaluated on VQA as a downstream task. ViLBERT is pre-trained using Conceptual Captions~\cite{Sharma2018ConceptualCA} and fine-tuned on \env. To evaluate how well a model trained on VQAv2 or OK-VQA performs on \env, we fine-tune ViLBERT after training them on thoese datasets. These models are referred to as `ViLBERT-VQA' and `ViLBERT-OK-VQA' in Table\ \textcolor{red}{5}.

\textbf{KRISP~\cite{Marino2021KRISPII}}
KRISP is a method for knowledge-based VQA which combines multi-modal Transformers with graph neural networks methods on knowledge graphs. We use the same models and data and knowledge sources and pre-processing steps as in that work, but filter the knowledge graph based on \env \ rather than OK-VQA (see Sec. 3.2 of~\cite{Marino2021KRISPII}).

\textbf{GPV-2~\cite{Kamath2022WeblySC}} GPV-2~\cite{Kamath2022WeblySC} is a generative vision and language model built using the T5~\cite{Raffel2020ExploringTL} language model and VinVL~\cite{Zhang2021VinVLRV} image features. It was pre-trained on Conceptual Captions~\cite{Sharma2018ConceptualCA} and then fine-tuned in a multi-task setting on image captioning, visual question answering, object localization, and classification, as well on web-search images for 10,000 visual concepts. 

We fine-tune the fully-trained model on \env\ by training it to generate the most common answer for each question. For direct answer evaluations, answers are then generated using beam search with 20 beams. For multiple choice, the answers are ranked by the log-probability score assigned to them by the model.

We perform two additional experiments with rationales with this model. First, ground-truth rationales are appended to the question as additional input text. Recall that we do not provide rationales at test time. However, for this experiment we use them during test. We refer to this model as `GPV-2 + GT Ratl.'. Second, we use the same setting, but we replace every occurrence of the ground-truth answer in the rationale with the $[$answer$]$ token. We refer to this model as `GPV-2~\cite{Kamath2022WeblySC} + Masked Ans.' in Table\ \textcolor{red}{5}.

\end{document}